\documentclass[letterpaper, 10pt, conference]{ieeeconf}
\IEEEoverridecommandlockouts

\usepackage{cite}
\usepackage{amsmath,amssymb,amsfonts}
\usepackage{algorithmic}
\usepackage{graphicx}
\usepackage{textcomp}
\usepackage{xcolor}
\usepackage{multirow}
\usepackage{hyperref}
\usepackage{fancyhdr}
\fancypagestyle{withfooter}{
  
  \fancyfoot[C]{\footnotesize Accepted to the IEEE ICRA Workshop on Field Robotics 2024}
}

\title{\LARGE \bf
A comparison between single-stage and two-stage 3D tracking algorithms for greenhouse robotics
}
\author{David Rapado-Rincon$^{1}$, Akshay K. Burusa$^{1}$, Eldert J. van Henten$^{1}$ and Gert Kootstra$^{1}$
\thanks{Research funded by the FlexCRAFT project (NWO grant p17-01)}
\thanks{$^{1}$ Farm Technology Group, Wageningen University \& Research, Wageningen, The Netherlands {\tt\small david.rapadorincon@wur.nl}}%
}

\begin{document}
\maketitle
\thispagestyle{withfooter}
\pagestyle{withfooter}

\setlength\floatsep{8pt}
\setlength\textfloatsep{8pt}

\begin{abstract}
With the current demand for automation in the agro-food industry, accurately detecting and localizing relevant objects in 3D is essential for successful robotic operations. However, this is a challenge due the presence of occlusions. Multi-view perception approaches allow robots to overcome occlusions, but a tracking component is needed to associate the objects detected by the robot over multiple viewpoints. Multi-object tracking (MOT) algorithms can be categorized between two-stage and single-stage methods. Two-stage methods tend to be simpler to adapt and implement to custom applications, while single-stage methods present a more complex end-to-end tracking method that can yield better results in occluded situations at the cost of more training data. The potential advantages of single-stage methods over two-stage methods depends on the complexity of the sequence of viewpoints that a robot needs to process. In this work, we compare a 3D two-stage MOT algorithm, 3D-SORT, against a 3D single-stage MOT algorithm, MOT-DETR, in three different types of sequences with varying levels of complexity. The sequences represent simpler and more complex motions that a robot arm can perform in a tomato greenhouse. Our experiments in a tomato greenhouse show that the single-stage algorithm consistently yields better tracking accuracy, especially in the more challenging sequences where objects are fully occluded or non-visible during several viewpoints.
\end{abstract}


\section{Introduction}
The agricultural and food sectors are increasingly under pressure from a rising global population and a concurrent decrease in available labor. Automation, particularly through the deployment of robotic technologies, has emerged as a solution in addressing these problems. However, the integration of robots into these sectors faces significant challenges, notably in the domain of robotic perception due to complex environmental conditions, such as occlusions and variation \cite{kootstra_selective_2021}.

An accurate and efficient representation of the robot's environment, including the relevant objects for a given task, is crucial for a successful robot operation in these environments \cite{crowley_dynamic_1985,elfring_semantic_2013}. Incorporating multiple views into a single representation have the potential to improve detection and localization even in highly occluded conditions \cite{arad_development_2020}. For this, active perception algorithms are key in selecting optimal viewpoints \cite{burusa_efficient_2023}. However, building representations from multi-view perception requires associating upcoming detections with their corresponding object representations and previous measurements \cite{elfring_semantic_2013, persson_semantic_2020, rapado-rincon_development_2023}. In object-centric representations, this task is often referred to as multi-object tracking (MOT). 

\begin{figure}[ht]
    \centering
    \includegraphics[width=0.45\textwidth]{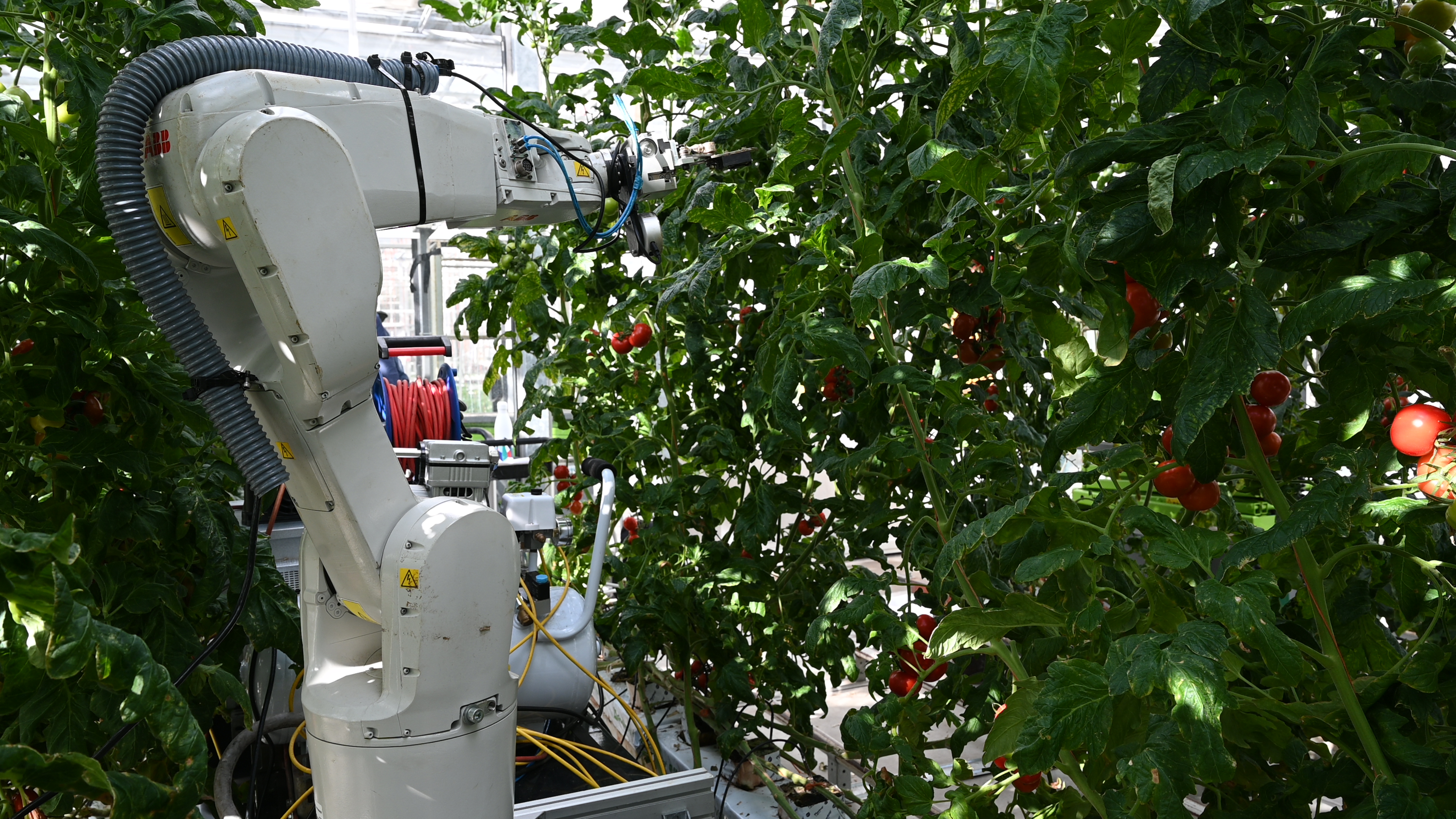}
    \caption{Robotic system used for data collection. The robot arm ABB IRB1200 is mounted over a mobile platform that allows motion over the greenhouse row rails. A scissor-like cutting and gripping tool, and a Realsense L515 camera are mounted on the end-effector of the robot.}
    \label{fig:robot_greenhouse}
\end{figure}

A common robotic system that can be used for plant monitoring, maintenance and harvesting is a robotic arm with both a camera and a gripper placed in the end effector, as shown in Figure \ref{fig:robot_greenhouse}. This robot could be tasked to harvest tomatoes in a greenhouse, which first requires detecting and localizing tomatoes in the plant. There are occlusions due to the tomatoes growing in trusses, and the presence of leaves from the target plant and nearby plants. Therefore, to localize and properly estimate properties such as the ripeness of the tomatoes, the robot needs to collect multiple viewpoints along the plant and track the detected tomatoes over all the viewpoints. In this situation, the accuracy of the tomato tracking algorithm will determine the accuracy of the plant representation and therefore, the ability of the robot to fulfill its task. 

Recent advancements in MOT algorithms have led to the development of various approaches, prominently categorized into two-stage and single-stage methods. Two-stage methods, exemplified by SORT \cite{bewley_simple_2016} and its more sophisticated successor, DeepSORT \cite{wojke_simple_2017}, separate the detection and re-identification (re-ID) feature extraction steps. In the first stage objects are detected using a deep learning model. In the second stage, re-ID properties such as object position for SORT and/or re-ID black box features like in DeepSORT are generated for every object detected in the first stage. Single-stage methods, such as FairMOT \cite{zhang_fairmot_2021}, use a deep learning algorithm to simultaneously perform object detection and re-ID feature extraction within a single network inference step. This reduces the computational requirements and improves performance. However, this comes at the cost of an increased algorithm complexity and a need for larger training datasets. In both two-stage and single-stage methods, the re-ID properties and features are used by a data association algorithm to match newly detected objects with those of previously tracked objects.

In agro-food environments, 2D two-stage MOT algorithms are the most commonly used approaches for tracking. They have been employed for tasks like crop monitoring and fruit counting \cite{halstead_fruit_2018, kirk_robust_2021, halstead_crop_2021, hu_lettucetrack_2022, villacres_apple_2023, rapado-rincon_development_2023, rapado-rincon_minksort_2023-1}. The increased complexity and need for larger training datasets of single-stage methods, combined with the novelty of these algorithms, results in a lower presence of these methods in applied agro-food environments. In our previous work, we developed MOT-DETR \cite{rapado-rincon_mot-detr_2024}, a 3D single-stage MOT algorithm, and showed that it outperformed state-of-the-art MOT algorithms like FairMOT for tracking and monitoring tomatoes in greenhouses.

In the tomato harvesting system described above and shown in Figure \ref{fig:robot_greenhouse}, the accuracy of MOT algorithms tends to decrease with the level of occlusion \cite{rapado-rincon_development_2023}. The amount of occlusion that the robot encounters in the sequence of viewpoints depends on two factors. The first being the amount of clutter in the environment, resulting in occluding leaves. The second factor relates to the motion of the robot. For example, if a robot arm moves at a low speed with the camera always pointing toward the same area of the environment, the resulting sequence will be similar to a video where the overlap between consecutive frames is large. However, due to obstacles in the environment and presence of occluding objects, a robot arm might have to perform motions where the camera is seeing a completely different area of the environment for a while. This results in sequences where objects might disappear from the field-of-view (FoV) of the camera for a few viewpoints, and where the difference between two consecutive viewpoint images is large. The resulting challenging sequences might require for more powerful and complex MOT algorithms, like single-stage ones. However, this might come at the cost of increased complexity and amount of training data.

The performance differences and trade-offs of two-stage and single-stage MOT algorithms has not been widely studied in agro-food robotic environments. Hu et al. \cite{hu_lettucetrack_2022} compared several 2D single-stage and two-stage algorithms in a lettuce field using a wheeled robot. The sequence of images generated by the camera system of the robot contain a high overlap between consecutive viewpoints or frames due the the 2D motion of the wheeled robot. They show how in this situation, two-stage algorithms perform better than single-stage ones. However, this conclusion might not be applicable to more complex scenes and 3D robot movements on arm-based robotic systems. 

This paper presents a comparison of a 3D two-stage MOT algorithm, 3D-sort \cite{rapado-rincon_development_2023}, with a single-stage one, MOT-DETR \cite{rapado-rincon_mot-detr_2024}, under different type of motions generated by a robot arm in a tomato greenhouse. The different motions represent different frame-to-frame distances and occlusions.

\section{Materials and Methods}

\subsection{Data}
\begin{figure}[ht]
    \centering
    \includegraphics[width=0.45\textwidth]{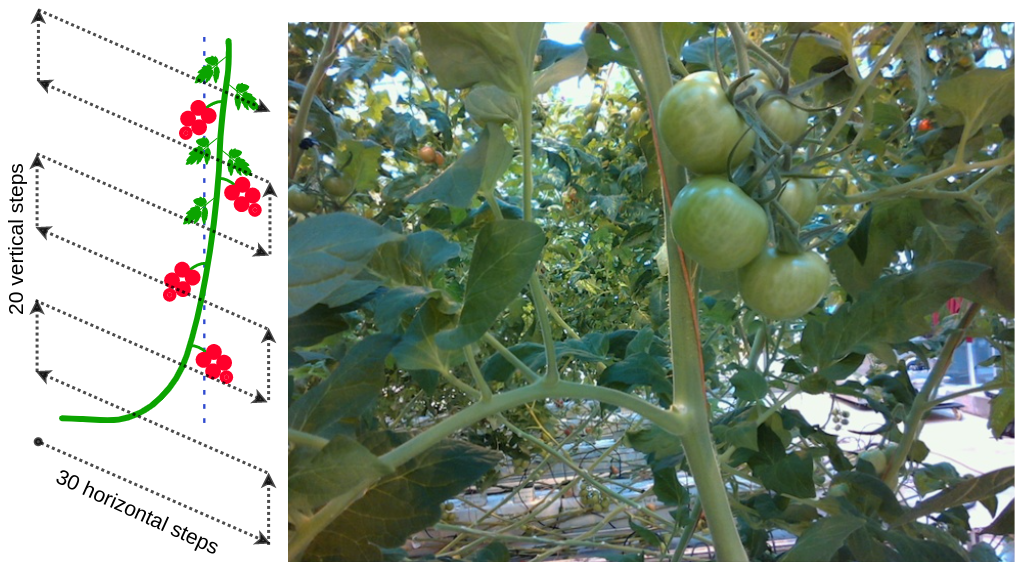}
    \caption{\textbf{Left.} Illustration of the path followed by the robot to collect viewpoints of real plants. \textbf{Right.} Example of a viewpoint on a plant.}
    \label{fig:real_plant}
\end{figure}

A dataset using five real plants from a tomato greenhouse was previously collected using the system shown in Fig. \ref{fig:robot_greenhouse} \cite{rapado-rincon_mot-detr_2024}. Per plant, viewpoints were collected using a planar motion sequence in front of each plant at a distance of 40 cm and 60 cm as shown in Fig. \ref{fig:real_plant}. In total, data from 5,400 viewpoints from five plants was collected. Each viewpoint resulted in a color image, a point cloud whose origin corresponds to the robot fixed coordinate frame, and the ground truth bounding boxes and tracking IDs of each tomato present in the viewpoint.

\begin{table}[htbp]
\centering
\caption{Distribution of plants and images in train, validation and test splits.}
\resizebox{\linewidth}{!}{
    \begin{tabular}{cccc}
    \hline
    \textbf{Type} & \textbf{Split} & \textbf{\# Plants} & \textbf{\# Viewpoints} \\ \hline
    \multirow{2}{*}{Real} & Train / Validation & 4 & 3,570 / 630 \\
    & Test & 1 & 1,200 \\ \hline
    \end{tabular}
}
\label{tab:dataset}
\end{table}

The data was divided into train, validation and test splits as shown in Table \ref{tab:dataset}. To prevent plants from being seen by the networks at train time and during the experiments, train and validation splits came from the same pool of four plants, while test splits were generated from a different plant.

\begin{figure*}[ht]
    \centering
    \includegraphics[width=\textwidth]{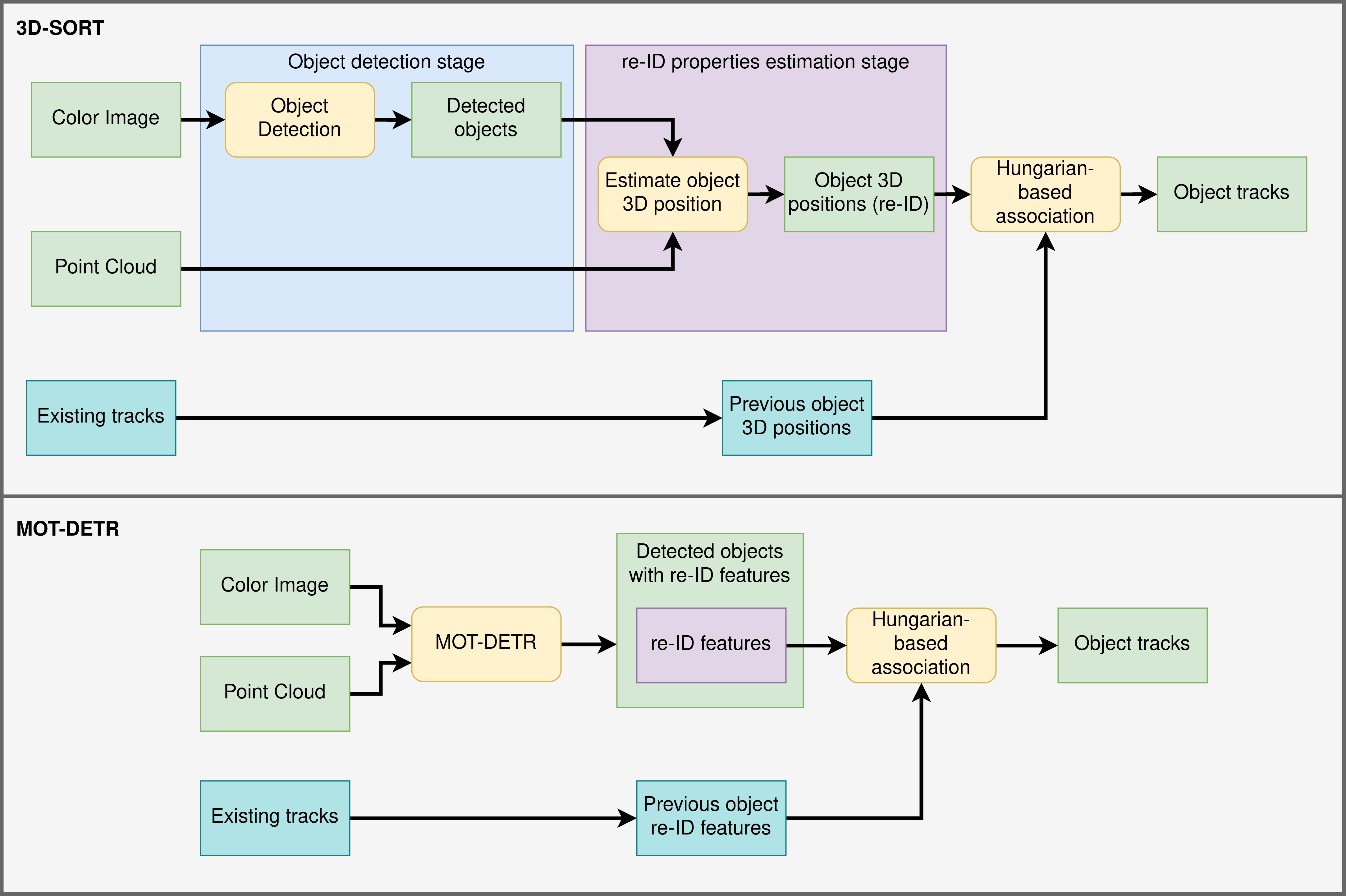}
    \caption{3D-SORT (top). First, the color image is processed by the object detection algorithm. The resulting detections are used together with the point cloud to generate a 3D position per detected object, which corresponds to the re-ID property used by the data association step. The Hungarian algorithm is then used to associate the locations of newly detected object with the previously tracked object positions. MOT-DETR (bottom). Color images and point clouds are used at the same time to detect objects with their corresponding class and re-ID features, which are black box features. The re-ID features are then passed to a Hungarian-based data association algorithm.}
    \label{fig:alg}
\end{figure*}

\subsection{Algorithms}
In this work, we compared two 3D MOT algorithms, 3D-SORT \cite{rapado-rincon_development_2023} and MOT-DETR \cite{rapado-rincon_mot-detr_2024}.  Both algorithms take as input the same data: a color image and its corresponding point cloud transformed into the robot's fixed coordinate system. 3D-SORT, shown in Figure \ref{fig:alg}-top, is a two-stage MOT algorithm, while MOT-DETR, shown in Figure \ref{fig:alg}-bottom, is a single-stage MOT algorithm.

3D-SORT is a two-stage MOT algorithm that at every viewpoint detects objects and estimates their 3D location given a color image and a point cloud. 3D-SORT starts with a detection algorithm that detects objects in a color image. In this work, this step is performed with a YOLOv8 \cite{noauthor_ultralyticsultralytics_nodate} network trained with the train set described in Table \ref{tab:dataset}. Then, each tomato bounding box is used to filter out the corresponding points from the point cloud. For each detected tomato, its corresponding points are used to estimate its 3D position with respect to the fixed robot coordinate frame. These 3D positions are used in the data association step to assign new detections with previously detected objects. The data association is performed using the Hungarian algorithm on a cost matrix calculated using the Mahalanobis distance between the location of newly detected tomatoes and the location of previously tracked tomatoes. A Kalman filter is then used to update the location of each tracked tomato at every viewpoint and association. For more details of 3D-SORT, we refer to its original paper \cite{rapado-rincon_development_2023}.

MOT-DETR is a single-stage MOT algorithm that uses the detection and tracking approach of FairMOT \cite{zhang_fairmot_2021} but with the detection architecture of DETR \cite{carion_end--end_2020}. Furthermore, it was extended to process simultaneously 2D images and 3D point clouds. For every viewpoint, MOT-DETR predicts the following outputs per detected object: 2D bounding box, class, and re-ID features. The data association is done by using the re-ID features of newly detected and previously detected objects to build a cost matrix using the cosine distance. This cost matrix is then passed to a Hungarian algorithm that generates the associations for tracking. We trained MOT-DETR as described in the original paper using the same data, which corresponds to the train set depicted in Table \ref{tab:dataset}. For more details, we refer to the original paper \cite{rapado-rincon_mot-detr_2024}.

\subsection{Experiments}
We evaluated 3D-SORT and MOT-DETR under different sequence types that represent different motions and occlusion levels that a robot arm might encounter:

\begin{itemize}
    \item \textbf{Sort.} For our test set, we had two sequences of 600 frames at two distances from the plant. For each sequence, we selected 100 random viewpoints. Then we sequentially ordered the selected viewpoints with the objective of minimizing frame-to-frame distance. This resulted in a 100 frame sequence similar to a video sequence with low frame rate. 
    \item \textbf{Random.} Similarly to the previous experiment, 100 viewpoints were selected out of the pool of viewpoints per sequence and plant. However, they were not ordered in this case. This resulted in a sequence where jumps between frames are larger, and objects were occluded during several frames.
    \item \textbf{AP.} Active Perception (AP) improves the capacity of robots to deal with occlusions in unknown environments like tomato greenhouses \cite{burusa_efficient_2023}. Therefore, we studied the performance of our algorithm when 100 viewpoints were selected out of the pool of viewpoints using the AP algorithm developed by \cite{burusa_efficient_2023} that maximizes information gain between viewpoints. This method requires the definition of a region-of-interest (RoI). For this experiment, the region of interest was defined as a rectangular volume, approximately set around the middle-front of the robot where the stem of the tomato plant was expected. 
\end{itemize}

We evaluated the tracking accuracy using the metric High Order Tracking Accuracy (HOTA), with its sub-metrics Localization Accuracy (LocA), Detection Accuracy (DetA) and Association Accuracy (AssA); and with Multi-Object Tracking Accuracy (MOTA) and its sub-metric ID Switches (IDSW). Each experiment was repeated five times by randomly selecting a different set of 100 random viewpoints.

\section{Results \& Discussion}
The results of our experiments can be seen in Table \ref{tab:results}. 3D-SORT surpasses MOT-DETR in DetA and LocA. This outcome was anticipated, given 3D-SORT's reliance on YOLOv8, a network pre-trained on large object detection datasets. In contrast to 3D-SORT, MOT-DETR's simultaneous task on object detection and re-identification with a single network compromises its detection efficacy, as networks tasked with multiple objectives typically underperform those dedicated to a single task \cite{zhang_fairmot_2021}.

In terms of overall tracking accuracy, as measured by HOTA and MOTA scores, alongside AssA and IDSW, MOT-DETR consistently outperformed 3D-SORT across all tests. This performance difference underscores the ability of MOT-DETR to better understand a scene and the relationships between objects in it. The re-ID features generated by MOT-DETR are generated by a network that has access to the whole color image and point cloud data for each viewpoint. This means that relationships between the objects can be encoded by the network to boost tracking performance. This is in contrast to two-stage methods like 3D-SORT where the re-ID properties and features are obtained from the object detection in 2D and/or 3D without considering any other information about the environment and nearby detections. 

When the sequence of viewpoints was generated by an Active Perception (AP) algorithm, the tracking performance of both 3D-SORT and MOT-DETR increases. While observing the sequences, the selected viewpoints have a lower level of occlusion in front of the plant stem, which is the region of interest of the AP algorithm. Furthermore, in the AP sequences the tomato trusses are always in the center areas of the image, in contrast to the partially visible trusses in both Sort and Random sequences. This setup significantly reduces the likelihood of trusses and tomatoes being partially visible, thus mitigating detection and tracking challenges. This outcome shows the advantageous impact of AP in enhancing tracking accuracy under conditions where occlusions are prevalent.

\begin{table}[htbp]
\centering
\caption{Tracking performance results on different type of viewpoint sequences.}
\resizebox{\linewidth}{!}{
    \begin{tabular}{cccccccc}
    \hline
    \textbf{Dataset} & \textbf{Algorithm} & \textbf{HOTA}$\uparrow$ & \textbf{DetA}$\uparrow$ & \textbf{AssA}$\uparrow$ & \textbf{LocA}$\uparrow$ & \textbf{MOTA}$\uparrow$ & \textbf{IDSW}$\downarrow$ \\ \hline
    \multirow{2}{*}{Sort} & 3D-SORT & 52.45 & \textbf{65.59} & 42.13 & \textbf{92.79} & 65.95 & 41.8 \\ 
    & MOT-DETR & \textbf{60.4} & 56.3 & \textbf{65.17} & 79.17 & \textbf{70.38} & \textbf{23.8} \\ \hline
    \multirow{2}{*}{Random} & 3D-SORT & 33.37 & \textbf{66.17} & 16.92 & \textbf{92.81} & 46.24 & 200.2 \\ 
    & MOT-DETR & \textbf{59.63} & 56.32 & \textbf{63.44} & 79.14 & \textbf{66.86} & \textbf{57.4} \\ \hline
    \multirow{2}{*}{AP} & 3D-SORT & 63.99 & \textbf{79.89} & 52.08 & \textbf{93.21} & 76.95 & 84.2 \\ 
    & MOT-DETR & \textbf{71.66} & 66.08 & \textbf{78.58} & 80.16 & \textbf{88.09} & \textbf{4.2} \\ \hline
    \end{tabular}
}
\label{tab:results}
\end{table}

\section{Conclusion}
In this work, we have compared a two-stage 3D MOT algorithm, 3D-SORT, against a single-stage MOT algorithm, MOT-DETR. We showed how the two-stage method 3D-SORT yields better object detection results due to a more powerful object detection algorithm. However, the single-stage method, MOT-DETR, is able to consistently outperform 3D-SORT in overall tracking and data association performance. This shows that even with lower detection performance, the single-stage method is able to better understand the scene and encode the objects and their relationships. The difference in performance is larger with more complex sequences where frame-to-frame distance is large, and objects might disappear from the camera FoV during several viewpoints. Furthermore, we showed how using active perception to reduce the number of occlusions present in the sequences boosts the tracking accuracy of both methods.

\bibliographystyle{IEEEtran}  
\bibliography{references}

\begin{thebibliography}{10}
\providecommand{\url}[1]{#1}
\csname url@rmstyle\endcsname
\providecommand{\newblock}{\relax}
\providecommand{\bibinfo}[2]{#2}
\providecommand\BIBentrySTDinterwordspacing{\spaceskip=0pt\relax}
\providecommand\BIBentryALTinterwordstretchfactor{4}
\providecommand\BIBentryALTinterwordspacing{\spaceskip=\fontdimen2\font plus
\BIBentryALTinterwordstretchfactor\fontdimen3\font minus \fontdimen4\font\relax}
\providecommand\BIBforeignlanguage[2]{{%
\expandafter\ifx\csname l@#1\endcsname\relax
\typeout{** WARNING: IEEEtran.bst: No hyphenation pattern has been}%
\typeout{** loaded for the language `#1'. Using the pattern for}%
\typeout{** the default language instead.}%
\else
\language=\csname l@#1\endcsname
\fi
#2}}

\bibitem{kootstra_selective_2021}
\BIBentryALTinterwordspacing
G.~Kootstra, X.~Wang, P.~M. Blok, J.~Hemming, and E.~van Henten, ``\BIBforeignlanguage{en}{Selective {Harvesting} {Robotics}: {Current} {Research}, {Trends}, and {Future} {Directions}},'' \emph{\BIBforeignlanguage{en}{Current Robotics Reports}}, vol.~2, no.~1, pp. 95--104, Mar. 2021. [Online]. Available: \url{https://doi.org/10.1007/s43154-020-00034-1}
\BIBentrySTDinterwordspacing

\bibitem{crowley_dynamic_1985}
\BIBentryALTinterwordspacing
J.~Crowley, ``\BIBforeignlanguage{en}{Dynamic world modeling for an intelligent mobile robot using a rotating ultra-sonic ranging device},'' in \emph{\BIBforeignlanguage{en}{Proceedings. 1985 {IEEE} {International} {Conference} on {Robotics} and {Automation}}}, vol.~2.\hskip 1em plus 0.5em minus 0.4em\relax St. Louis, MO, USA: Institute of Electrical and Electronics Engineers, 1985, pp. 128--135. [Online]. Available: \url{http://ieeexplore.ieee.org/document/1087380/}
\BIBentrySTDinterwordspacing

\bibitem{elfring_semantic_2013}
\BIBentryALTinterwordspacing
J.~Elfring, S.~van~den Dries, M.~van~de Molengraft, and M.~Steinbuch, ``\BIBforeignlanguage{en}{Semantic world modeling using probabilistic multiple hypothesis anchoring},'' \emph{\BIBforeignlanguage{en}{Robotics and Autonomous Systems}}, vol.~61, no.~2, pp. 95--105, Feb. 2013. [Online]. Available: \url{https://linkinghub.elsevier.com/retrieve/pii/S0921889012002163}
\BIBentrySTDinterwordspacing

\bibitem{arad_development_2020}
\BIBentryALTinterwordspacing
B.~Arad, J.~Balendonck, R.~Barth, O.~Ben‐Shahar, Y.~Edan, T.~Hellström, J.~Hemming, P.~Kurtser, O.~Ringdahl, T.~Tielen, and B.~v. Tuijl, ``\BIBforeignlanguage{en}{Development of a sweet pepper harvesting robot},'' \emph{\BIBforeignlanguage{en}{Journal of Field Robotics}}, vol. n/a, no. n/a, 2020, \_eprint: https://onlinelibrary.wiley.com/doi/pdf/10.1002/rob.21937. [Online]. Available: \url{https://www.onlinelibrary.wiley.com/doi/abs/10.1002/rob.21937}
\BIBentrySTDinterwordspacing

\bibitem{burusa_efficient_2023}
\BIBentryALTinterwordspacing
A.~K. Burusa, J.~Scholten, D.~R. Rincon, X.~Wang, E.~J. van Henten, and G.~Kootstra, ``Efficient {Search} and {Detection} of {Relevant} {Plant} {Parts} using {Semantics}-{Aware} {Active} {Vision},'' June 2023, arXiv:2306.09801 [cs]. [Online]. Available: \url{http://arxiv.org/abs/2306.09801}
\BIBentrySTDinterwordspacing

\bibitem{persson_semantic_2020}
\BIBentryALTinterwordspacing
A.~Persson, P.~Z.~D. Martires, A.~Loutfi, and L.~De~Raedt, ``\BIBforeignlanguage{en}{Semantic {Relational} {Object} {Tracking}},'' \emph{\BIBforeignlanguage{en}{IEEE Transactions on Cognitive and Developmental Systems}}, vol.~12, no.~1, pp. 84--97, Mar. 2020, arXiv: 1902.09937. [Online]. Available: \url{http://arxiv.org/abs/1902.09937}
\BIBentrySTDinterwordspacing

\bibitem{rapado-rincon_development_2023}
\BIBentryALTinterwordspacing
D.~Rapado-Rincón, E.~J. van Henten, and G.~Kootstra, ``\BIBforeignlanguage{en}{Development and evaluation of automated localisation and reconstruction of all fruits on tomato plants in a greenhouse based on multi-view perception and {3D} multi-object tracking},'' \emph{\BIBforeignlanguage{en}{Biosystems Engineering}}, vol. 231, pp. 78--91, July 2023. [Online]. Available: \url{https://www.sciencedirect.com/science/article/pii/S1537511023001162}
\BIBentrySTDinterwordspacing

\bibitem{bewley_simple_2016}
\BIBentryALTinterwordspacing
A.~Bewley, Z.~Ge, L.~Ott, F.~Ramos, and B.~Upcroft, ``\BIBforeignlanguage{en}{Simple {Online} and {Realtime} {Tracking}},'' \emph{\BIBforeignlanguage{en}{2016 IEEE International Conference on Image Processing (ICIP)}}, pp. 3464--3468, Sept. 2016, arXiv: 1602.00763. [Online]. Available: \url{http://arxiv.org/abs/1602.00763}
\BIBentrySTDinterwordspacing

\bibitem{wojke_simple_2017}
N.~Wojke, A.~Bewley, and D.~Paulus, ``Simple online and realtime tracking with a deep association metric,'' in \emph{2017 {IEEE} {International} {Conference} on {Image} {Processing} ({ICIP})}, Sept. 2017, pp. 3645--3649, iSSN: 2381-8549.

\bibitem{zhang_fairmot_2021}
\BIBentryALTinterwordspacing
Y.~Zhang, C.~Wang, X.~Wang, W.~Zeng, and W.~Liu, ``\BIBforeignlanguage{en}{{FairMOT}: {On} the {Fairness} of {Detection} and {Re}-identification in {Multiple} {Object} {Tracking}},'' \emph{\BIBforeignlanguage{en}{International Journal of Computer Vision}}, vol. 129, no.~11, pp. 3069--3087, Nov. 2021. [Online]. Available: \url{https://doi.org/10.1007/s11263-021-01513-4}
\BIBentrySTDinterwordspacing

\bibitem{halstead_fruit_2018}
\BIBentryALTinterwordspacing
M.~Halstead, C.~McCool, S.~Denman, T.~Perez, and C.~Fookes, ``\BIBforeignlanguage{en}{Fruit {Quantity} and {Ripeness} {Estimation} {Using} a {Robotic} {Vision} {System}},'' \emph{\BIBforeignlanguage{en}{IEEE Robotics and Automation Letters}}, vol.~3, no.~4, pp. 2995--3002, Oct. 2018. [Online]. Available: \url{https://ieeexplore.ieee.org/document/8392450/}
\BIBentrySTDinterwordspacing

\bibitem{kirk_robust_2021}
R.~Kirk, M.~Mangan, and G.~Cielniak, ``\BIBforeignlanguage{en}{Robust {Counting} of {Soft} {Fruit} {Through} {Occlusions} with {Re}-identification},'' in \emph{\BIBforeignlanguage{en}{Computer {Vision} {Systems}}}, ser. Lecture {Notes} in {Computer} {Science}, M.~Vincze, T.~Patten, H.~I. Christensen, L.~Nalpantidis, and M.~Liu, Eds.\hskip 1em plus 0.5em minus 0.4em\relax Cham: Springer International Publishing, 2021, pp. 211--222.

\bibitem{halstead_crop_2021}
\BIBentryALTinterwordspacing
M.~Halstead, A.~Ahmadi, C.~Smitt, O.~Schmittmann, and C.~McCool, ``Crop {Agnostic} {Monitoring} {Driven} by {Deep} {Learning},'' \emph{Frontiers in Plant Science}, vol.~12, 2021. [Online]. Available: \url{https://www.frontiersin.org/article/10.3389/fpls.2021.786702}
\BIBentrySTDinterwordspacing

\bibitem{hu_lettucetrack_2022}
\BIBentryALTinterwordspacing
N.~Hu, D.~Su, S.~Wang, P.~Nyamsuren, and Y.~Qiao, ``\BIBforeignlanguage{English}{{LettuceTrack}: {Detection} and tracking of lettuce for robotic precision spray in agriculture},'' \emph{\BIBforeignlanguage{English}{Frontiers in Plant Science}}, vol.~13, Sept. 2022, publisher: Frontiers. [Online]. Available: \url{https://www.frontiersin.org/journals/plant-science/articles/10.3389/fpls.2022.1003243/full}
\BIBentrySTDinterwordspacing

\bibitem{villacres_apple_2023}
\BIBentryALTinterwordspacing
J.~Villacrés, M.~Viscaino, J.~Delpiano, S.~Vougioukas, and F.~Auat~Cheein, ``\BIBforeignlanguage{en}{Apple orchard production estimation using deep learning strategies: {A} comparison of tracking-by-detection algorithms},'' \emph{\BIBforeignlanguage{en}{Computers and Electronics in Agriculture}}, vol. 204, p. 107513, Jan. 2023. [Online]. Available: \url{https://www.sciencedirect.com/science/article/pii/S0168169922008213}
\BIBentrySTDinterwordspacing

\bibitem{rapado-rincon_minksort_2023-1}
\BIBentryALTinterwordspacing
D.~Rapado-Rincón, E.~J. van Henten, and G.~Kootstra, ``{MinkSORT}: {A} {3D} deep feature extractor using sparse convolutions to improve {3D} multi-object tracking in greenhouse tomato plants,'' July 2023, arXiv:2307.05219 [cs]. [Online]. Available: \url{http://arxiv.org/abs/2307.05219}
\BIBentrySTDinterwordspacing

\bibitem{rapado-rincon_mot-detr_2024}
\BIBentryALTinterwordspacing
D.~Rapado-Rincon, H.~Nap, K.~Smolenova, E.~J. van Henten, and G.~Kootstra, ``{MOT}-{DETR}: {3D} {Single} {Shot} {Detection} and {Tracking} with {Transformers} to build {3D} representations for {Agro}-{Food} {Robots},'' Feb. 2024, arXiv:2311.15674 [cs]. [Online]. Available: \url{http://arxiv.org/abs/2311.15674}
\BIBentrySTDinterwordspacing

\bibitem{noauthor_ultralyticsultralytics_nodate}
\BIBentryALTinterwordspacing
``ultralytics/ultralytics: {NEW} - {YOLOv8} in {PyTorch} {\textgreater} {ONNX} {\textgreater} {OpenVINO} {\textgreater} {CoreML} {\textgreater} {TFLite}.'' [Online]. Available: \url{https://github.com/ultralytics/ultralytics}
\BIBentrySTDinterwordspacing

\bibitem{carion_end--end_2020}
\BIBentryALTinterwordspacing
N.~Carion, F.~Massa, G.~Synnaeve, N.~Usunier, A.~Kirillov, and S.~Zagoruyko, ``End-to-{End} {Object} {Detection} with {Transformers},'' May 2020, arXiv:2005.12872 [cs]. [Online]. Available: \url{http://arxiv.org/abs/2005.12872}
\BIBentrySTDinterwordspacing

\end{thebibliography}

\end{document}